\documentclass{egpubl_sca_poster}
\usepackage{sca2018p}

\WsPaper         
 \electronicVersion 

\ifpdf \usepackage[pdftex]{graphicx} \pdfcompresslevel=9
\else \usepackage[dvips]{graphicx} \fi

\PrintedOrElectronic

\usepackage{t1enc,dfadobe}

\usepackage{egweblnk}
\usepackage{cite}
\usepackage{multirow}
\usepackage[labelfont=bf, textfont=it]{caption}
\title[Dilated Temporal Fully-Convolutional Network]%
      {Dilated Temporal Fully-Convolutional Network for Semantic Segmentation of Motion Capture Data}

\author[Noshaba Cheema \& Somayeh Hosseini]
{\parbox{\textwidth}{\centering Noshaba Cheema$^{1,2,3}$,
        Somayeh Hosseini$^{1,2}$\thanks{First two authors contributed equally; email: ncheema@mpi-inf.mpg.de}, Janis Sprenger$^{1,2}$, Erik Herrmann$^{1,2}$, Han Du$^{1,2}$, Klaus Fischer$^{1}$ and Philipp Slusallek$^{1,2}$
        }
        \\
{\parbox{\textwidth}{\centering $^1$DFKI Saarbr{\"u}cken, $^2$Saarland University, $^3$Max-Planck Institute for Informatics; Germany
       }
}
}

\begin{document}

\teaser{
\centering
\includegraphics[width=0.90\linewidth]{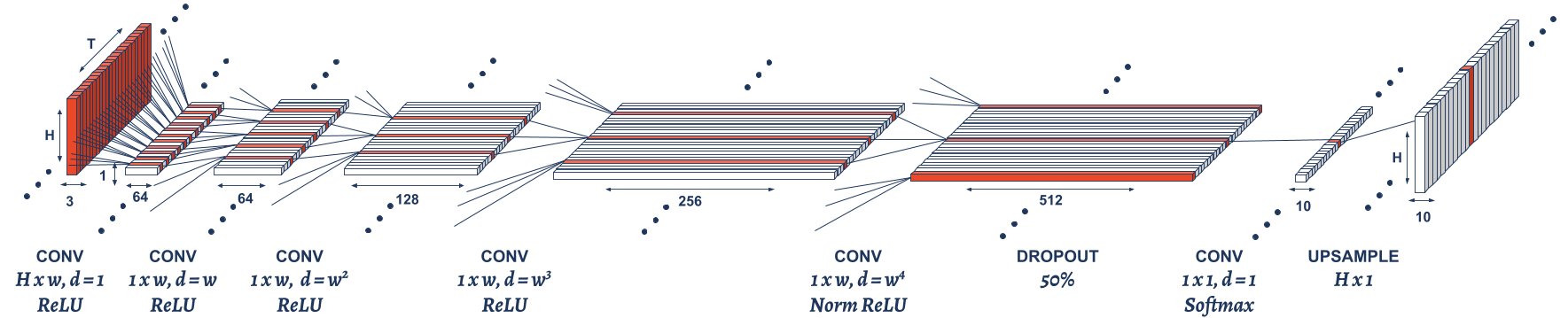}
\caption{Our dilated temporal fully-convolutional neural network (DTFCN) for motion capture segmentation. The initial layer consists of a traditional 2D convolutional layer. The next layers are 1D temporal acausal convolutions with dilation. The final dilated conv layer uses a normalizing ReLU activation function. Finally, we add a Softmax layer and upsample the output. \\}
\label{fig:tfcn}
}

\maketitle

\begin{abstract}
Semantic segmentation of motion capture sequences plays a key part in many data-driven motion synthesis frameworks. It is a preprocessing step in which long recordings of motion capture sequences are partitioned into smaller segments. Afterwards, additional methods like statistical modeling can be applied to each group of structurally-similar segments to learn an abstract motion manifold. The segmentation task however often remains a manual task, which increases the effort and cost of generating large-scale motion databases. We therefore propose an automatic framework for semantic segmentation of motion capture data using a dilated temporal fully-convolutional network. Our model outperforms a state-of-the-art model in action segmentation, as well as three networks for sequence modeling. We further show our model is robust against high noisy training labels.


\end{abstract} 
\section{Our Proposed Architecture}
Recurrent neural networks (RNN) are the go-to method to model time-dependent sequences. However, one of their major drawbacks is the exploding and vanishing gradient problem and the difficulty to parallelize their training. Additionally, \cite{TCN2018} have shown that temporal convolutional networks (TCN) perform just as well or even better than RNNs in sequence modeling tasks. Hence, we introduce a model, which is inspired by traditional image segmentation approaches \cite{long2015fully} and recent advances in sequence modeling \cite{TCN2018} for semantic
segmentation of motion capture data. 

\begin{figure}
\centering
\includegraphics[width=0.6\linewidth]{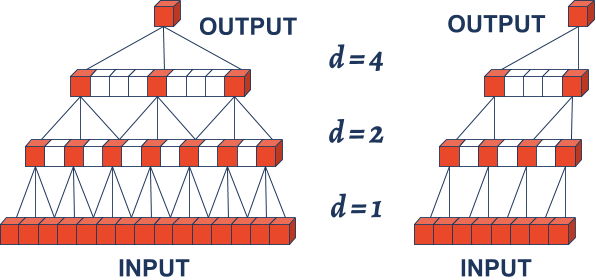}
\caption{Dilated convolution. Left: acausal dilation. Right: causal dilation. Systematic dilation increases the receptive field size exponentially.\\[-1cm]}
\label{fig:dilation}
\end{figure}

\begin{figure}
\centering
\includegraphics[width=0.8\linewidth]{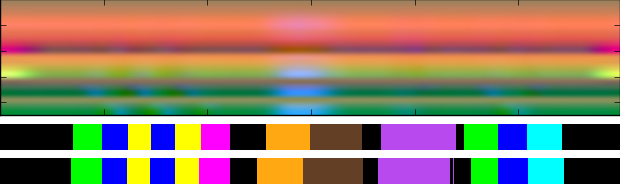}
\caption{Top: Motion capture sequence in RGB image domain. Middle: True labels. Bottom: Our predictions.\\}
\label{fig:prediction}
\end{figure}

In a preprocessing step, we first transform our motion capture data to an RGB image domain, much in the spirit of \cite{laraba20173d}. Each column of the image represents a frame in the motion sequence. 
The rows represent the joints and the RGB values are the scaled XYZ Euclidean coordinates of each corresponding joint. Such a \emph{motion image} can be seen in Fig. \ref{fig:prediction} (top). We then pass it to our network (Fig. \ref{fig:tfcn}). Akin to the five areas in our visual cortex (V1 - V5) \cite{Remington2012233}, our model has a total of five convolution layers. The initial layer consists of a traditional 2D convolutional layer which is only applied in the time dimension. To do that, we set the kernel height to the height of the image. Every layer has the same convolution width $w$ with stride 1. The next four layers are 1D temporal acausal convolutions with dilation. The dilation rate $d$ increases with each layer $l$, according to $d = w^{l-1}$. A convolutionized dense layer with a Softmax activation function is added after that. We found that a normalizing ReLU function \cite{Lea2017TemporalCN} before the Softmax layer increases accuracy.


Fig. \ref{fig:dilation} shows how dilated convolutions increase the receptive field exponentially without loss of resolution for acausal and causal convolutions. Causal convolutions are used for temporal data, where the output depends on previous samples only. Since our goal is to distinguish motions like \emph{left step} (step while walking) from \emph{begin} and \emph{end left step} (step from/to standing position), we use acausal convolutions, as these motion types rely on past and future information.

\section{Experiments and Results}

\begin{figure}
\includegraphics[width=\linewidth]{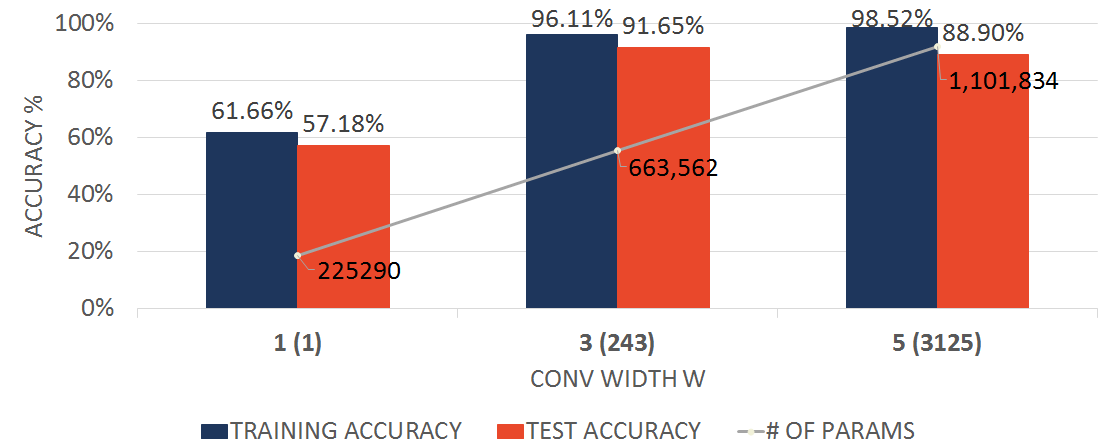}
\caption{Number of parameters (line) vs. train (dark blue) and test (red) accuracy for different convolution widths. The receptive field size with dilations after the first five layers is written in parentheses.\\}
\label{fig:recep}
\end{figure}

\begin{table}
\centering
\begin{tabular}{ |l|c|c|c|c|c| }
  \hline
  & ED-TCN & WaveNet & TDNN & LSTM & Ours\\
  \hline
  Train & 90.05\% & 90.64\% & 87.22\% & 86.32\% & \textbf{96.11\%}\\
  Test & 88.69\% & 88.47\% & 85.54\% & 81.95\% & \textbf{91.64\%}\\
  \hline
\end{tabular}
\caption{Comparison against other models. ED-TCN: \cite{Lea2017TemporalCN}, WaveNet: \cite{wavenet}, TDNN: \cite{tdnn}, LSTM: \cite{lstm}}
\label{tab:results}
\end{table}

Our motion capture dataset consists of  70 sequences with 10 motion labels: \emph{standing}, \emph{left/right step}, \emph{begin/end left step}, \emph{begin/end right step}, \emph{reach}, \emph{retrieve} and \emph{turn}. Our sequences reach up to 1500 frames. In all of our experiments, we use non-randomized 7-fold cross-validation. We use the Adam \cite{kingma2014adam} optimizer with 100 epochs for training. 

\begin{figure}
\centering
\includegraphics[width=0.8\linewidth]{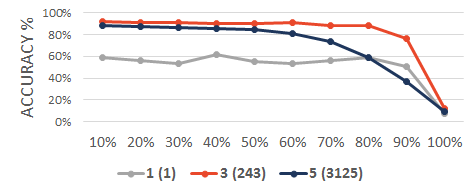}
\caption{Test accuracies for different receptive field sizes depending on noise level.}
\label{fig:noise}
\end{figure}

In order to determine the optimal receptive field size (RFS), we test our model on different convolution kernel widths $w$. Fig. \ref{fig:recep} shows that even though a width $w = 5$ (RFS: 3125 frames) covers the entire sequence, the accuracy does not differ much from using $w = 3$ (RFS: 342 frames). A width $w = 3$ uses $\sim$438K fewer parameters, however. Since our model has to be robust against human error due to wrongly-classified labels, we further train our model on noisy labels and test it on the true labels. Fig. \ref{fig:noise} shows despite adding 80\% noisy labels in the training data, an accuracy of over 88\% is reached on the true test labels for $w = 3$.

We test our model ($w = 3$) against another state-of-the-art TCN model \cite{Lea2017TemporalCN} for action segmentation and three commonly used neural network models \cite{wavenet, lstm, tdnn} for sequence modeling and classification using our dataset without noisy labels, and show that our model is superior to these models (Tab. \ref{tab:results}).

With this work, we have shown that our model provides a fruitful segmentation tool for motion capture segmentation. To support various types of motion, we further plan on increasing our motion image database and do more experiments on noisy data.

\small{\section*{Acknowledgements}
This work is funded by the European Union's Horizon 2020 research and innovation program under the Marie Sklodowska-Curie grant agreement No 642841; and by the German Federal Ministry of Education and Research (BMBF) through the project Hybr-iT under the grant 01IS16026A.}
\bibliographystyle{eg-alpha-doi}
\bibliography{sections/References}

\end{document}